\documentclass[runningheads]{llncs}
\usepackage{graphicx}
\usepackage{cite}
\usepackage{booktabs}
\usepackage{makecell}
\usepackage{multirow}
\usepackage{amsmath,amssymb,amsfonts}
\usepackage{color}
\begin{document}
	\title{ProtoDiv: Prototype-guided Division of Consistent Pseudo-bags for Whole-slide Image Classification}
	\titlerunning{ProtoDiv}
	%
	\author{Rui Yang \and Pei Liu \and Luping Ji \thanks{Corresponding author: L. Ji (\email{jiluping@uestc.edu.cn}).}}
	\institute{School of Computer Science and Engineering, University of Electronic Science and Technology of China, Chengdu, Sichuan, China}
	\maketitle
	\begin{abstract}
Due to the limitations of inadequate Whole-Slide Image (WSI) samples with weak labels, pseudo-bag-based multiple instance learning (MIL) appears as a vibrant prospect in WSI classification. However, the pseudo-bag dividing scheme, often crucial for classification performance, is still an open topic worth exploring.
 Therefore, this paper proposes a novel scheme, ProtoDiv, using a bag prototype to guide the division of WSI pseudo-bags. Rather than designing complex network architecture, this scheme takes a plugin-and-play approach to safely augment WSI data for effective training while preserving sample consistency. Furthermore, we specially devise an attention-based prototype that could be optimized dynamically in training to adapt to a classification task. We apply our ProtoDiv scheme on seven baseline models, and then carry out a group of comparison experiments on two public WSI datasets. Experiments confirm our ProtoDiv could usually bring obvious performance improvements to WSI classification.
		
	\end{abstract}
	
	\section{Introduction}
	With the promotion of whole-slide scanning technology, digital slides have been widely used in pathology diagnosis \cite{Cooper2012,hanna2022integrating}. The virtual slide system with automatic microscope produces digital Whole Slide Images (WSIs) by scanning and seamlessly splicing traditional glass slides \cite{kumar2020}. This kind of high-quality digital image, often viewed as a gold standard for cancer diagnosis, provides a chance for computer-aided diagnosis tools \cite{bera2019artificial,niazi2019digital,baxi2022digital}. However, a single WSI usually has an extremely-high resolution (20,000$\times$20,000 pixels) and massive microscopic bioinformation, posing great challenges to conventional image analysis. 
	
	Owing to the considerable advances in computer vision, many attempts have been made to train deep learning models on WSI data for computer-aided pathology diagnosis \cite{Li2020dsmil,lu2021,laleh2022benchmarking}. Among them, the weakly-supervised framework, based on embedding-level multiple instance learning (MIL) \cite{Carbonneau2018,ilse2018attn}, has been extensively studied in recent years. 
	Upon this frame, each WSI is tiled into a bag of sub-image patches with a bag-level weak annotation for training embedding-level MIL models.
	A multitude of methods with this framework focuses on exploiting mutual instance information using neural networks like RNN \cite{Campanella2019}, GCN \cite{Guan_2022_CVPR, Hou2022H2MIL, Zhao2020}, or Transformer \cite{Shao2021TransMIL, Chen2022HIPT, Li2021DTMIL}. They are proved to perform better in bag-level classification \cite{WANG2018Revisiting,ilse2018attn} when compared to other patch-based ones \cite{MIML2017, Campanella2019, Hou2016patchBasedCNN} using weak patch labels and patch-based neural networks. 
	
	Nonetheless, most of these existing works have not yet paid enough attention to the problem inherent in WSI data, namely, available WSIs are still at a small-scale \cite{liu2022advmil} while each one is made up of 10k $\sim$ 100k patches. This problem is believed to cause overfitting in WSI models, as stated in \cite{Zhang_2022_CVPR}. To alleviate it, two pioneering works, BDOCOX \cite{Shao2021BDOCOX} and DTFD-MIL \cite{Zhang_2022_CVPR}, first propose to generate pseudo-bags as training samples from an individual WSI bag for classification. The former attempts to generate sub-bags by sampling from the patch clusters indicating heterogeneous tissue patterns. 
	And the latter obtains pseudo-bags by randomly dividing WSI bags. Despite their success, two problems remain unstudied for the new paradigm of \textit{pseudo-bag-based MIL}. \textbf{(1)} Could this pseudo-bag-based paradigm be utilized as a generalized tool for MIL frameworks? Further exploration of the effect of pseudo-bags on MIL frameworks is needed, as previous works only take pseudo-bag generation as a component of their methods. \textbf{(2)} How to better divide pseudo-bags for WSI classification. We argue that existing schemes, \textit{e.g.}, random dividing and cluster-based dividing, could yield the pseudo-bags with a lack of consistency or adaptability. Particularly, the random dividing, a simple scheme ignoring the intrinsic tissue phenotype in WSIs, is likely to break the consistency between a parent bag and its child pseudo-bags. Thus, the child pseudo-bags could not represent their parent bag well, thereby imposing noisy samples on models. As for cluster-based dividing, it has taken tissue phenotype into consideration but is extremely time-consuming for clustering vast patches. Moreover, both schemes generate the pseudo-bags fixed in iterative model training, making pseudo-bags impossible to be adjusted adaptively. 
	
\begin{figure}[tp]
		\centering
		\includegraphics[width=0.99\textwidth]{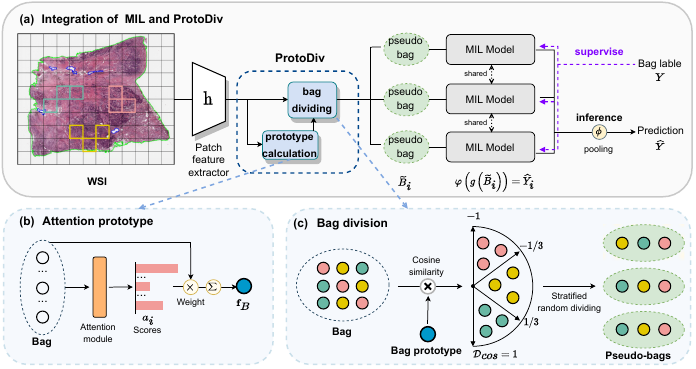}
		\caption{An overview of ProtoDiv. One WSI is used as an example to explain how our ProtoDiv works in MIL framework: extract patch features, calculate bag prototype, divide bag into pseudo-bags for training, and aggregate to get a final prediction.}
		\label{fig:framework}
	\end{figure}

	Accordingly, this paper studies the new paradigm of pseudo-bag-based MIL for WSI classification and proposes a simple yet efficient dividing method. This method, named ProtoDiv (see Fig. \ref{fig:framework}), introduces a bag prototype to guide bag dividing to generate consistent pseudo-bags as training samples. Specifically, we derive a bag prototype, a mean-based static one, or an attention-based adaptive one, from each WSI at first. Based on a normalized metric of the similarity between this prototype and each instance, we divide all instances to generate pseudo-bags by stratified random sampling. As a result, all pseudo-bags could have a consistent distribution and well represent their parent bag, in the context of tissue phenotype. With our ProtoDiv, MIL models could be trained on the scalable WSI data augmented by consistent pseudo-bags, thus alleviating the problems posed by limited WSI data. 
	
	The key contributions of this paper are summarized as follows. \textbf{(1)} This paper proposes a general and easy-to-use method (ProtoDiv), consisting of prototype calculation and bag dividing, to generate prototype-based consistent pseudo-bags for training MIL models. It is a plugin-and-play method for WSI data augmentation, often easily applied to most existing MIL methods. \textbf{(2)} We conduct extensive experiments on seven baseline models and two WSI datasets to validate our scheme. Empirical results demonstrate that our proposed scheme could often bring obvious performance improvements to existing MIL models. Moreover, ablation studies confirm the superiority of our prototype-based scheme over both random and cluster-based dividing. 
	
	\section{Proposed Methodology}
        In this section, we first formulate the pseudo-bags-based MIL paradigm with our ProtoDiv (Fig. \ref{fig:framework}(a)). Next, we present the two key components of ProtoDiv: prototype calculation (Fig. \ref{fig:framework}(b)) and prototype-based bag dividing (Fig. \ref{fig:framework}(c)).  
	\subsection{Problem Formulation}
	\subsubsection{Traditional MIL Settings.} 
	Given a WSI, we denote it by a bag of instances $B=\left \{ \left (\mathbf{f}_{1},y_{1}\right),\dots,\left(\mathbf{f}_{k},y_{k}\right),\dots,\left(\mathbf{f}_{K},y_{K}\right)\right \}$, where $K$ is the patch number of $B$,  $\mathbf{f}_{k}\in \mathbb{R}^{d}$ is the feature of the $k$-th instance with a corresponding label $ y_{k}$. Notably, $y_{k}$ is usually unavailable in MIL settings. In binary classification, $Y$ is positive only if $B$ contains at least one positive instance. Most MIL methods obtain the bag-level prediction via an instance aggregator $g$ and a simple bag-level classifier $\varphi$, formulated as $\widehat{Y}=\varphi\left (g(\mathbf{f}_{1},\dots,\mathbf{f}_{k},\dots,\mathbf{f}_{K})\right)$, \textit{i.e.}, $\widehat{Y}=\varphi\left ( g(B)\right)$. The bags used as training samples under this setting are often from a limited number of WSIs, which could cause overfitting\cite{Zhang_2022_CVPR} and impede model performance.
	\subsubsection{Integrating MIL with ProtoDiv.} 
	To augment bag data, we divide $B$ into $n$ pseudo-bags with ProtoDiv, \textit{i.e.}, $B=\{ \widetilde{B}_{1},\dots,\widetilde{B}_{i},\dots,\widetilde{B}_{n}\}$. The instance set of each pseudo-bag $\widetilde{B}_{i}$ is disjoint with others and has the same label as its parent bag $B$. Namely, the label of $\widetilde{B}_{i}$ is $Y_{i}=Y$. In training, all pseudo-bags are viewed as individual samples to train a MIL model. Hence, the prediction of $\widetilde{B}_{i}$ is  $\widehat{Y}_{i}=\varphi\left(g\left(\widetilde{B}_{i}\right)\right)$, and the classification loss for training pseudo-bags is
   \begin{equation}
  \mathcal{L}= -\frac{1}{n}\sum^{n}_{i}Y_i\log{\widehat{Y}_{i}} +\left ( 1-Y_i \right ) \log{\left ( 1-\widehat{Y}_{i} \right ) }. 
  \label{eq1}
  \end{equation}
   While in inference, a bag-level prediction is collected from the prediction of all pseudo-bags, written as
    \begin{equation}\widehat{Y}=\phi \left(\widehat{Y}_{1},\dots,\widehat{Y}_{i},\dots,\widehat{Y}_{n}\right),\end{equation}
	where $\phi$ is a pooling operator that aggregates pseudo-bag-level predictions. 
	
    \subsection{Our ProtoDiv Scheme}
    \subsubsection{Prototype Calculation.}
    We introduce bag prototype for pseudo-bag generating, based on two observations: i) phenotype-based dividing could well preserve the consistency between each pseudo-bag and its parent bag; ii) heuristic cluster algorithms like K-means are often time-expensive although considering tissue phenotype. 
    Specifically, we propose a simple and intuitive strategy to calculate a bag prototype $\mathbf{f}_{B}\in \mathbb{R}^{d}$, a vector with the same dimension as each instance. This prototype is $\mathbf{f}_{B}=\frac{1}{K} \sum_{k}^{K} \mathbf{f}_{k}$, called a \textit{mean-based} prototype. It assumes that the mean of all the instance features in $B$ could well represent $B$. 

    However, this mean-based prototype is static and could not adapt to a specific downstream task. Therefore, we further propose an \textit{attention-based} prototype, as shown in Fig. \ref{fig:framework}(b).  
    Concretely, a simple and computation-efficient gated-attention module \cite{ilse2018attn} is utilized to obtain this attention-based prototype. For the $i$-th instance of $B$, gated-attention yields its attention score as follows:
	\begin{equation}
		a_{k}=\frac{\exp \left\{\mathbf{w}^{\top}\left(\tanh \left(\mathbf{V} \mathbf{f}_{k}^{\top}\right) \odot \operatorname{sigm}\left(\mathbf{U} \mathbf{f}_{k} ^ { \top }\right)\right)\right\}}{\sum_{j=1}^{K} \exp \left\{\mathbf{w}^{\top}\left(\tanh \left(\mathbf{V} \mathbf{f}_{j}^{\top}\right) \odot \operatorname{sigm}\left(\mathbf{U} \mathbf{f}_{j}^{\top}\right)\right)\right\}}\in [0,1],
	\end{equation}
	where $\mathbf{w}$, $\mathbf{V}$, and $\mathbf{U}$ are trainable weight matrices. We obtain the attention prototype by: 
	\begin{equation}
		\mathbf{f}_{B}=\sum_{k}^{K} a_{k}\mathbf{f}_{k}.
	\end{equation}
    At each training epoch, the module, used for our attention-based prototype, is optimized before the pseudo-bag-level MIL model, by an independent classification loss. As a result, our attention-based prototype could well adapt to the classification task. Ablation studies also show the effectiveness of our attention-based prototype. Compared to a mean-based prototype, the attention-based one enjoys a dynamic and adaptive division of pseudo-bags during iterative training. Both two prototypes will be evaluated and discussed in experiments.
	
	\subsubsection{Bag Dividing via Prototype.}
	After patch feature extraction, we use the prototype calculated above to guide bag dividing (Fig. \ref{fig:framework}(c)). 
	Specifically, at first, we measure the cosine distance between each instance and $\mathbf{f}_{B}$ as follows:
	\begin{equation}
		\mathcal{D}_{cos}^{i}=\frac{\mathbf{f}_{B}\cdot \mathbf{f}_i}{||\mathbf{f}_{B}||\ ||\mathbf{f}_i||}\in [-1, 1].
	\end{equation}
	Based on this normalized metric indicating instance similarities, all instances are split into $l$ disjoint sections, each corresponding to a non-overlapping interval with a length of $2/l$ in cosine space. In this manner, each section could be identified as a patch cluster representing a specific tissue phenotype. 
	Then, all the patches within each section are split into $n$ parts randomly and evenly, and these $n$ parts are assigned to $n$ pseudo-bags one by one. Hence, we obtain $\{ \widetilde{B}_{1},\dots,\widetilde{B}_{i},\dots,\widetilde{B}_{n}\}$ from $B$, where $\widetilde{B}_{i}$ exactly consists of the $l$ parts with different phenotype, \textit{i.e.}, $\widetilde{B}_{i}$ could retain all tissue phenotypes presented in $B$. Note that when $l=1$, ProtoDiv would degenerate to a simple random dividing scheme that ignores tissue phenotype. 
 
    The proposed ProtoDiv not only could enable pseudo-bags to better represent their parent bag, but also maintains the consistency between pseudo-bags. Moreover, it contains an adaptive bag prototype so that pseudo-bags can be adjusted dynamically for better prediction. Consequently, ProtoDiv could be employed to boost the performance of MIL models by safely increasing the size of WSI data without imposing noisy samples. 
	\section{Experiment and Result}
	
	\subsubsection{Experimental Settings.} 
	The proposed method is evaluated on two WSI datasets from The
	Cancer Genome Atlas \cite{Kandoth2013} (TCGA): TCGA Lung Cancer dataset (TCGA-Lung) and TCGA Breast Cancer dataset (TCGA-BRCA) for tumor subtyping. (1) TCGA-Lung contains 1054 slides of two lung cancer subtypes, Lung Adenocarcinoma (LUAD) and Lung Squamous Cell Carcinoma (LUSC). (2) TCGA-BRCA dataset totally contains 1055 slides. After selecting out the slides with two invasive cancer subtypes, Invasive Ductal (IDC) and Invasive Lobular Carcinoma (ILC), there are 965 WSIs left. 
	
	In experiments, 4-fold cross-validation is adopted with a ratio of 60:15:25 for training, validation, and test sets. Performance is assessed on the test set by the best model on the validation set. We report area under curve (AUC) as the main evaluation index with sensitivity (Sen.) and accuracy (Acc.) also considered. 
	
	\subsubsection{Implementation Details.} 
	Following CLAM \cite{lu2021}, in pre-processing, tissue regions are segmented by a binary threshold method, and then the non-overlapping patches in a size of $256\times256$ are tiled from these regions at 10$\times$ magnification. Patches are fed into a pretrained truncated ResNet50 model from \cite{lu2021} for feature extraction. $\phi$ is implemented by a mean pooling function. All experiments are conducted in an environment of torch 1.9 and CUDA 11.1 on a machine with 2 GPUs of NVIDIA GeForce RTX 3080 Ti (12G).
\subsubsection{Classification Performance.} 
	We apply ProtoDiv to seven existing state-of-the-art models \cite{Shao2021TransMIL,ilse2018attn,Campanella2019,lu2021,Li2020dsmil,Zhang_2022_CVPR}. The results summarized in Table \ref{tab1} verify that ProtoDiv could often bring obvious improvements in performance to these baseline models. It suggests that ProtoDiv could be utilized as an effective data augmentation strategy to increase the size of WSI data so as to alleviate the problem of overfitting that easily occurs in WSI-based models. 
	\begin{table}[t]
		\small
		\centering
		\caption{Performance comparison between the baseline model with and without our ProtoDiv scheme. An attention-based prototype is used in ProtoDiv.}
		\label{tab1}
		\begin{tabular}{p{4cm}<{\centering}p{1cm}<{\centering}p{1.3cm}<{\centering}p{1.3cm}<{\centering}p{1.3cm}<{\centering}p{1.3cm}<{\centering}p{1.3cm}<{\centering}} 
			\toprule
			\multicolumn{1}{c}{\multirow{2}{*}{Classification Models}} & \multicolumn{3}{c}{TCGA-BRCA} & \multicolumn{3}{c}{TCGA-Lung} \\ 
			\cmidrule(lr){2-4} \cmidrule(lr){5-7}
			\multicolumn{1}{c}{}                        & AUC & Acc. & Sen. & AUC & Acc. & Sen.         \\ 
			\midrule
			ABMIL \cite{ilse2018attn}                   & 0.797   &  0.772    &  0.839         &  0.895	  &  0.843    &  0.846                \\
			ABMIL~+~\textbf{ProtoDiv}                         &  \textbf{0.863}   &  \textbf{0.799}    &  \textbf{0.860}  &  \textbf{0.920}   &  \textbf{	0.874}    &  \textbf{0.877}                      \\ 
			\cmidrule(lr){1-1}
			RNN-MIL \cite{Campanella2019}               &  0.747   & 0.690     & 0.773    & 0.885	& 0.810	& 0.794             \\
			RNN-MIL~+~\textbf{ProtoDiv}                      & \textbf{0.815}     & \textbf{0.893}      & \textbf{0.848}  & \textbf{0.921}     & \textbf{0.861}      & \textbf{0.858}                      \\ 
			\cmidrule(lr){1-1}
			CLAM-SB \cite{lu2021}                       & 0.833    &  0.719    &  	0.854 & 0.905	& 0.826	& 0.834               \\
			CLAM-SB~+~\textbf{ProtoDiv}                         & \textbf{0.864}     & \textbf{0.807}      & \textbf{	0.866}  &\textbf{0.930}     &\textbf{0.872}      &\textbf{0.874}           \\ 
			\cmidrule(lr){1-1}
			CLAM-MB \cite{lu2021}                  & 0.838    &  	0.716    &  0.868   & 0.915	& 0.842	& 0.847            \\
			CLAM-MB~+~\textbf{ProtoDiv}                    &\textbf{0.877}     &\textbf{0.819}      & \textbf{0.875}        &\textbf{0.938}     &\textbf{0.881}      &\textbf{0.882}           \\ 
			\cmidrule(lr){1-1}
			DS-MIL \cite{Li2020dsmil}                  & 0.778   &  	0.687    &  0.755   & 0.817	& 0.769	& 0.785            \\
			DS-MIL~+~\textbf{ProtoDiv}                    &\textbf{0.851}     &\textbf{0.790}      & \textbf{0.856}        &\textbf{0.920}     &\textbf{0.851}      &\textbf{0.858}           \\ 
			\cmidrule(lr){1-1}
			TransMIL \cite{Shao2021TransMIL}          & 0.860	& 0.807	& 0.867 & 0.934	& 0.874	& 0.878     \\
			TransMIL~+~\textbf{ProtoDiv}                       &\textbf{0.891}     &\textbf{0.845}      & \textbf{0.897}    &\textbf{0.942}     &\textbf{0.879}      &\textbf{0.883}             \\ 
			\cmidrule(lr){1-1}
			DTFD-MIL \cite{Zhang_2022_CVPR}             &0.845	&0.831	&0.869  &0.933	&0.881	&0.881  
			\\
			DTFD-MIL~+~\textbf{ProtoDiv}            &\textbf{0.861}	&\textbf{0.832}	&\textbf{0.887}  &\textbf{0.944}	&\textbf{0.885}	&\textbf{0.885} \\
			\bottomrule
		\end{tabular}
	\end{table}
 
	In detail, we have two experimental findings. (1) On TCGA-BRCA, ProtoDiv increases the AUC performance of seven different models by 1.6\% $\sim$ 7.3\%. And the AUC performance on TCGA-Lung also shows improvements of 0.8\% $\sim$ 10.3\%. These empirical facts reveal that ProtoDiv could be used as a general and effective method for WSI classification. (2) With ProtoDiv, even some basic MIL models, \textit{e.g.}, ABMIL, could achieve comparable performance with recent SOTA MIL models. Concretely, on TCGA-BRCA, the AUC of ABMIL + ProtoDiv is 0.863, which is higher than that of two original CLAM models (0.833 and 0.837). It also reaches the same performance as the original TransMIL (0.860), a stronger and more complex one among existing MIL models. This observation suggests that data augmentation could also be an effective and promising strategy in WSI modeling apart from improving the model itself. 
	
\subsubsection{Ablation Study.} 
	We study our prototype-based pseudo-bag dividing scheme through adding its key ingredients step by step.
     Besides, we measure the total time of dividing 100 randomly-chosen WSIs for efficiency comparisons. From Table \ref{tab2}, we have the following observations. (1) On 3 popular MIL models, the ProtoDiv with a better prototype is always superior to both random dividing and K-means dividing. 
	(2) Only for ABMIL, the mean-based prototype performs better than the attention-based one. One possible reason is that a static prototype may be more suitable for basic models. 
	(3) ProtoDiv greatly shortens the time compared with K-means dividing, and it pays only \textit{a little time cost} for performance improvements compared with random dividing.
	\begin{table}[tp]
		\centering
		\scriptsize
		\caption{Ablation study on our ProtoDiv, conducted on TCGA-BRCA, including the time cost of different division schemes on 100 randomly-sampled WSIs.}
		\label{tab2}
		\begin{tabular}{p{1.8cm}<{\centering}p{1.3cm}<{\centering}p{1.3cm}<{\centering}p{1.5cm}<{\centering}p{1.3cm}<{\centering}p{1.3cm}<{\centering}p{1.5cm}<{\centering}p{1.1cm}<{\centering}} 
			\toprule
			\multirow{2}{*}{Scheme} &\multirow{2}{*}{Phenotype} &\multirow{2}{*}{Prototype}&\multirow{2}{*}{Adaptability}&\multicolumn{3}{c}{AUC Performance} &\multirow{2}{*}{time (s)} \\ 
			\cmidrule(lr){5-7}
			&&&& ABMIL & TransMIL   & DTFD-MIL             \\ 
			\midrule
			Random  &&&&   0.861    &    0.883      &     0.845      &0.985          \\ 
			K-means  &\checkmark&&&   0.858    &    0.880      &     0.848      &144.231        \\ \cmidrule(lr){1-1}
			\textbf{ProtoDiv}$_\text{mean}$ &\checkmark&\checkmark&&  \textbf{0.872}     &   0.877      &   0.855   &1.296            \\
			\textbf{ProtoDiv}$_\text{attn}$ &\checkmark&\checkmark&\checkmark&  0.863     &   \textbf{0.891}       & \textbf{0.861}   &1.381                \\
			\bottomrule
		\end{tabular}
	\end{table}	 
 \begin{table}[tp]
		\centering
		\scriptsize
		\caption{AUC performance under different settings of $n$ (pseudo-bag number) and $l$ (tissue phenotype number).}
		\label{tab3}
		\begin{tabular}{p{1.1cm}<{\centering}p{1.1cm}<{\centering}p{1.1cm}<{\centering}p{1.1cm}<{\centering}p{1.1cm}<{\centering}p{1.1cm}<{\centering}p{1.1cm}<{\centering}p{1.1cm}<{\centering}p{1.1cm}<{\centering}p{1.1cm}<{\centering}} 
			\toprule
			\multirow{2}{*}{$l$} & \multicolumn{9}{c}{AUC Performance (at different value of $n$)} \\ 
			\cmidrule(lr){2-10}&1   & 3      & 6      & 8      & 10     & 15     & 20     & 25     & 30     \\ 
			\midrule			
			4  &\multirow{5}{*}{$0.7975$}& 0.8618 & 0.8458 & 0.8651 & 0.8651 & 0.8637 & 0.8655 & 0.8649 & 0.8667 \\
			6  && 0.8369 & 0.8602 & 0.8600 & 0.8600 & 0.8667 & 0.8674 & 0.8651 & 0.8673 \\
			8  && 0.8548 & 0.8594 & 0.8615 & 0.8640 & 0.8667 & 0.8648 & 0.8645 & 0.8678 \\
                12 && 0.8459 & 0.8585 & 0.8626 & 0.8629 & 0.8661 & 0.8694 & 0.8668 & 0.8682 \\
			15 && 0.8597 & 0.8421 & 0.8715 & 0.8689 & 0.8640 & 0.8630 & 0.8646 & 0.8611           \\
			\bottomrule
		\end{tabular}
	\end{table}

	Furthermore, we conduct experiments on TCGA-BRCA to study the effect of hyper-parameters ($n$ and $l$), in which ABMIL and its better mean-based prototype are used. In Tabel \ref{tab2},
	we have some notable findings as follows. (1) As the number of pseudo-bags increases (when $n\geq 10$), AUC becomes insensitive to $l$ while when $n$ is small, different values of $l$ are more likely to cause obvious fluctuation. (2) AUC performance ranges from $0.8369$ to $0.8715$ when using ProtoDiv, whereas it still surpasses the baseline model apparently ($0.7975 \text{ when } n=1$). 
  \subsubsection{Result Visualization.}
\begin{figure}[h]
\centering
		\begin{minipage}[t]{0.42\linewidth}
			\centering
			\includegraphics[width=\textwidth]{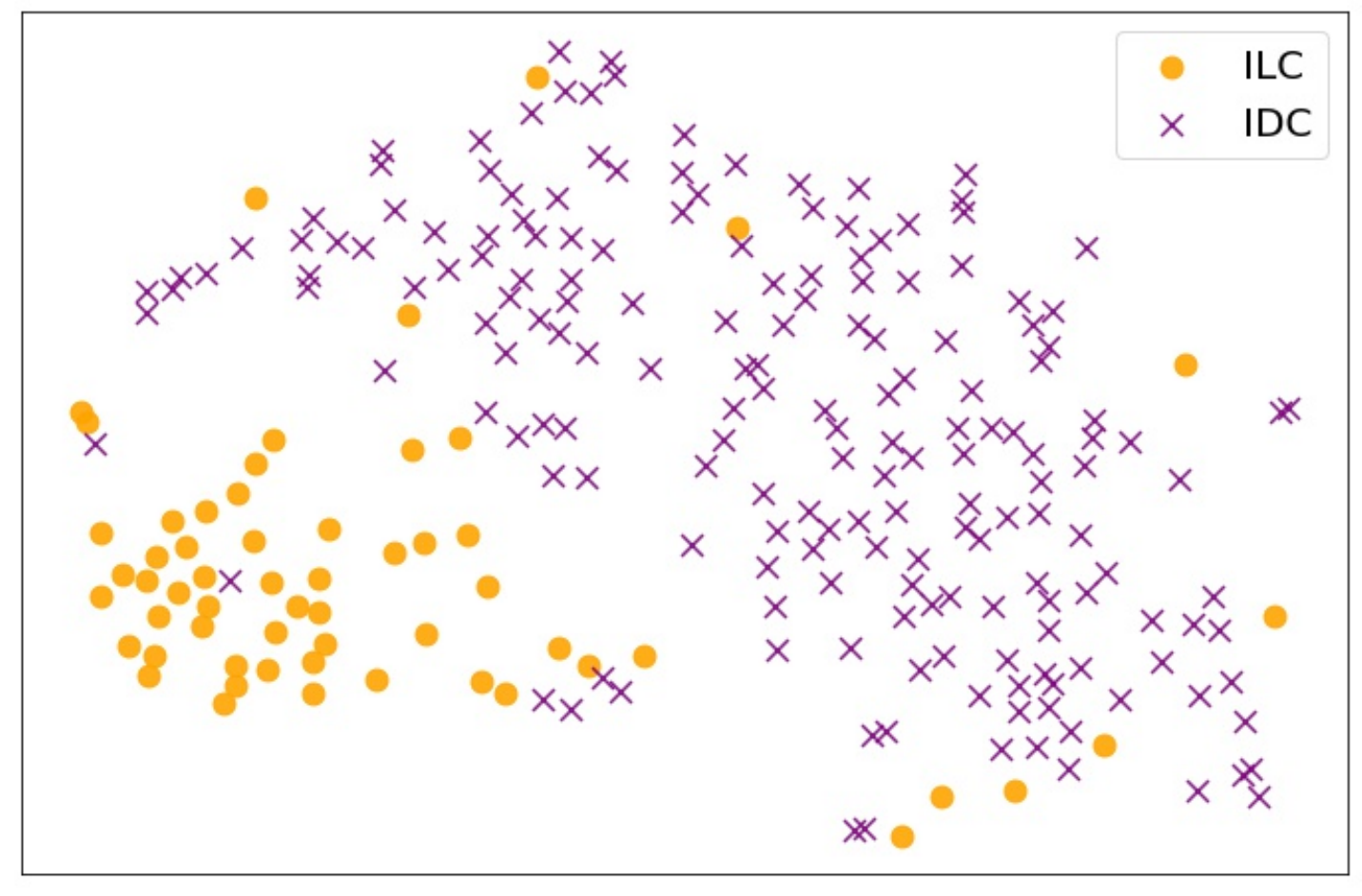}
		\end{minipage}
		\hspace{0.04\linewidth}
		\begin{minipage}[t]{0.42\linewidth}
			\centering
			\includegraphics[width=\textwidth]{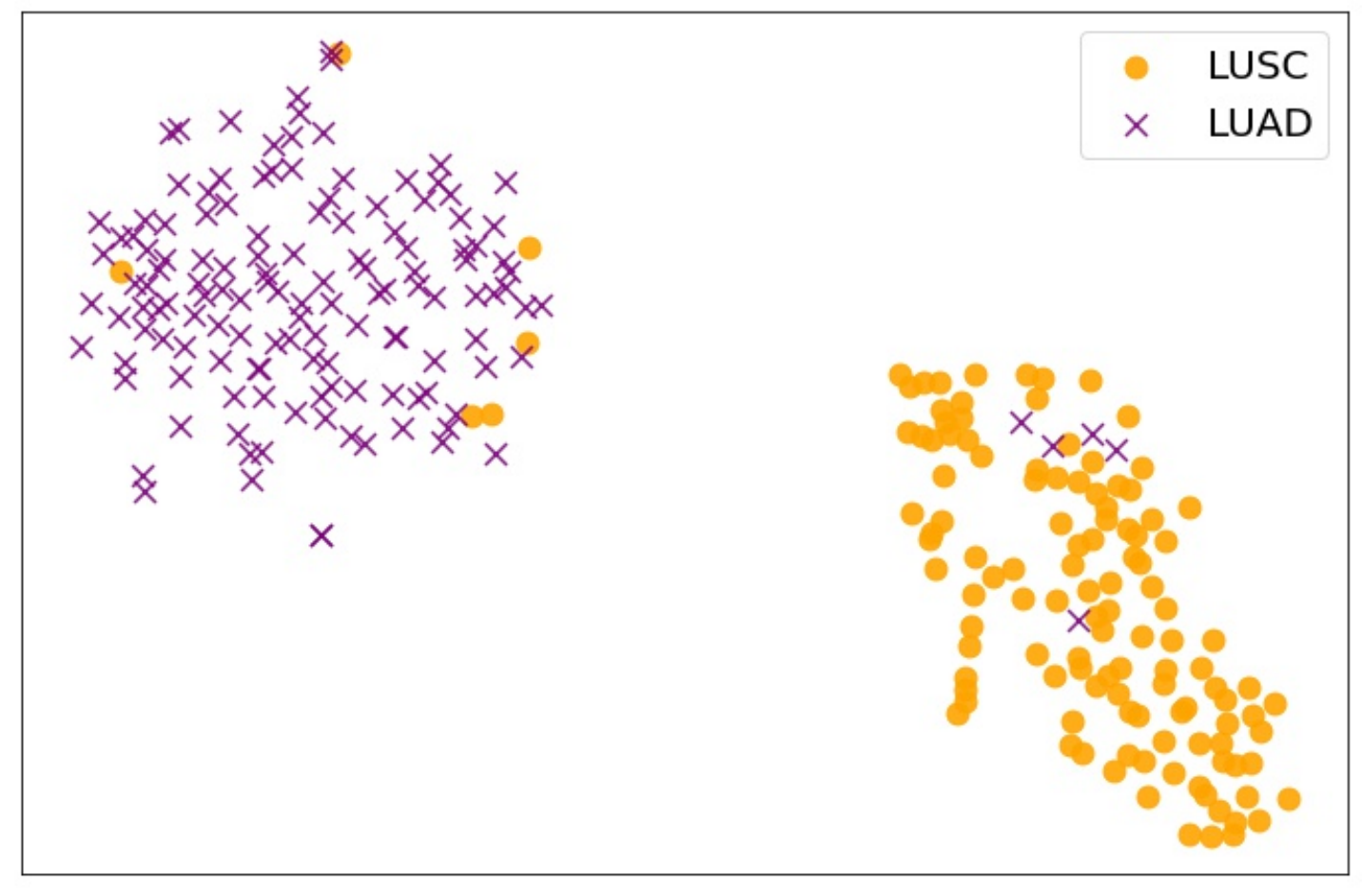}
		\end{minipage}
		\caption{Visualization of the attention prototypes in TCGA-BRCA (left) and TCGA-Lung (right) test set by using t-SNE \cite{vandermaaten08a}. ABMIL is adopted as the baseline model.}
		\label{fig:attnprototype}
	\end{figure}

 We visualize final attention-based prototypes in Fig. \ref{fig:attnprototype}. It implies that attention-based prototypes could well discriminate the cancer subtype of WSIs, \textit{i.e.}, this kind of prototype could well represent WSI bag and improve the quality of pseudo-bag dividing. As a result, the pseudo-bags generated by our attention prototype could be adaptive to classification task. 
		\begin{figure}[h]
		\centering		\includegraphics[width=0.92\textwidth]{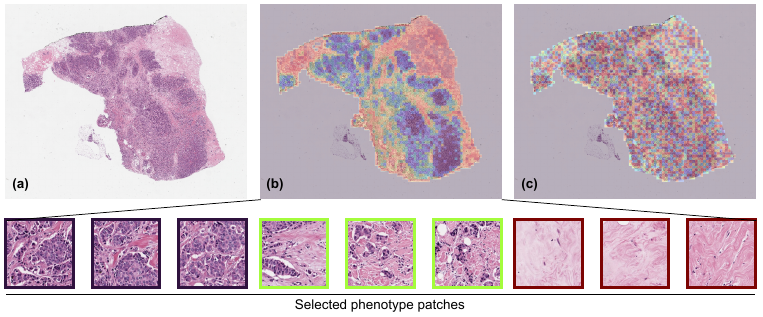}
		\caption{Visualization of the phenotype and pseudo-bag generated by our ProtoDiv. (a) is the original WSI. (b) shows the cluster prototypes via ProtoDiv. (c) gives the pseudo-bags generated via ProtoDiv. Each color indicates one specific phenotype or pseudo-bag in (b) and (c). More results are shown in our supplementary material.}
		\label{fig:visprotodiv}
	\end{figure}

We further visualize the phenotypes and pseudo-bags generated by our ProtoDiv in Fig. \ref{fig:visprotodiv}. In Fig. \ref{fig:visprotodiv}(b), the patches from different similarity sections are marked with different colors. Intuitively, they are shown to be clustered into different tissue phenotypes. In Fig. \ref{fig:visprotodiv}(c), we mark different pseudo-bags with different colors. It implies that each pseudo-bag comprises uniformly-dispersed patches. More visualization results are available in our supplementary material.
\subsubsection{Discussion}
To mitigate data inadequacy, we focus on augmenting the WSI data instead of designing complex network architectures. This study still has some limitations, \textit{e.g.}, the limited baseline MIL models and WSI datasets for comparisons, and a preliminary exploration of the pooling function $\phi(\cdot)$ used for aggregating pseudo-bag-level predictions. 
	\section{Conclusions}
	This paper proposes ProtoDiv to generate consistent pseudo-bags. It is a general and easy-to-use method that can be applied to most existing MIL models. Experiments demonstrate that our ProtoDiv could often help to improve the performance of existing MIL models. Moreover, visualization results indicate that ProtoDiv could retain the phenotype distribution so as to safely augment WSI data by consistent pseudo-bag dividing.
	%
	%
	%
	\bibliographystyle{splncs04}
	\bibliography{MyRefs}

\clearpage
\section*{Supplementary Material}

\begin{figure}[htbp]
\centering
\includegraphics[width=0.75\textwidth]{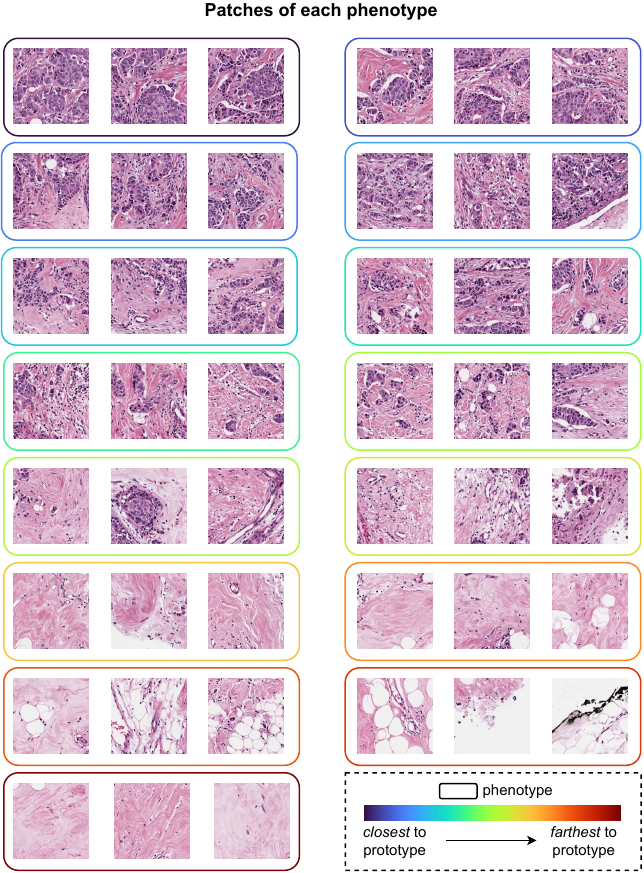}
\caption{All phenotypes and their respective patches, as a supplement to the Figure 3 in our main context. Patches are clustered into different tissue phenotypes for subsequent pseudo-bag generation.}
\label{sup-fig:visphenotype}
\end{figure}

\begin{figure}[htbp]
\centering
\includegraphics[width=\textwidth]{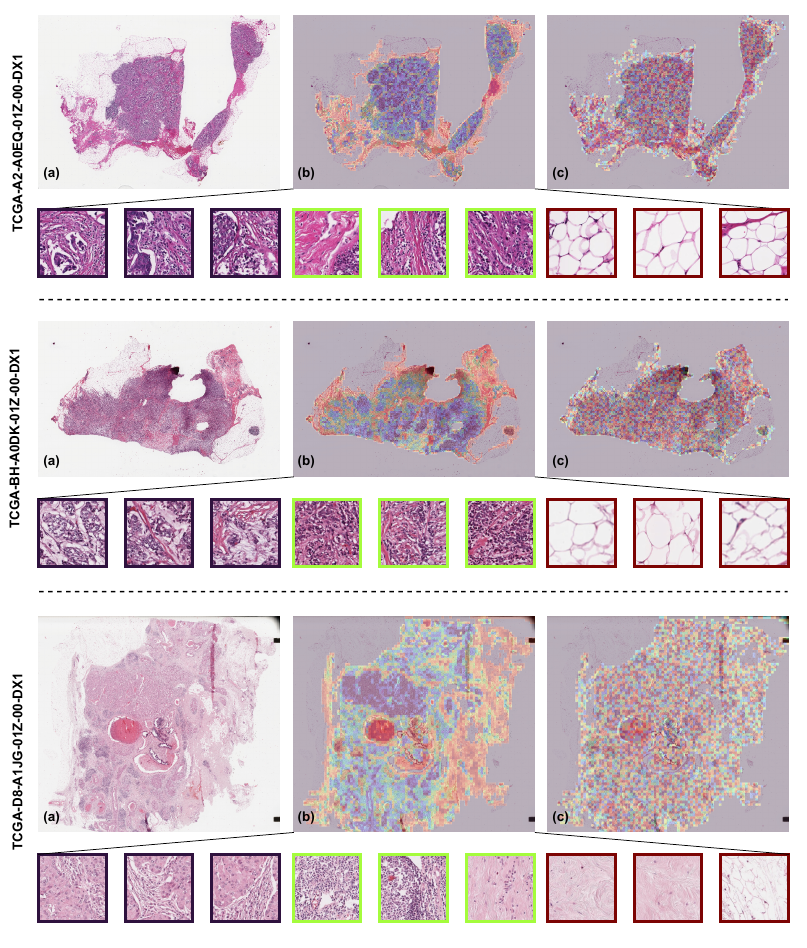}
\caption{Visualization of the phenotype and pseudo-bag generated by our ProtoDiv. (a) is the original WSI. (b) shows the cluster prototypes via ProtoDiv. (c) gives the pseudo-bags generated via ProtoDiv. Each color indicates one specific phenotype or pseudo-bag in (b) and (c), respectively. The patches of three selected phenotypes are also shown at the bottom of each row.}
\label{sup-fig:visprotodiv}
\end{figure}

\end{document}